\documentclass[review]{elsarticle}

\usepackage{lineno,hyperref}
\usepackage[linesnumbered,lined,boxed,commentsnumbered]{algorithm2e}
\usepackage[noend]{algpseudocode}
\usepackage{graphicx}
\usepackage{array}
\usepackage{multirow}
\usepackage{amsfonts}
\usepackage{amsmath, bm}
\usepackage{xcolor}
\usepackage{booktabs}
\usepackage{textcomp}
\usepackage{scrextend}
\usepackage{subfig}
\modulolinenumbers[1]

\makeatletter
\providecommand{\doi}[1]{%
  \begingroup
    \let\bibinfo\@secondoftwo
    \urlstyle{rm}%
    \href{http://dx.doi.org/#1}{%
      doi:\discretionary{}{}{}%
      \nolinkurl{#1}%
    }%
  \endgroup
}
\makeatother

\makeatletter
\def\ps@pprintTitle{%
   \let\@oddhead\@empty
   \let\@evenhead\@empty
   \def\@oddfoot{\reset@font\hfil\thepage\hfil}
   \let\@evenfoot\@oddfoot
}
\makeatother

\begin{document}

\begin{frontmatter}

\title{Causality Extraction Based on Self-Attentive BiLSTM-CRF with Transferred Embeddings}

\author{Zhaoning Li}
\ead{lizhn7@mail2.sysu.edu.cn}
\author{Qi Li}
\ead{liqi38@mail2.sysu.edu.cn}
\author{Xiaotian Zou}
\ead{zouxt5@mail2.sysu.edu.cn}
\author{Jiangtao Ren\corref{mycorrespondingauthor}}
\ead{issrjt@mail.sysu.edu.cn}
\address{School of Data and Computer Science, Sun Yat-Sen University, Guangzhou, Guangdong, P.R.China 510006}
\cortext[mycorrespondingauthor]{Corresponding author}

\begin{abstract}
Causality extraction from natural language texts is a challenging open problem in artificial intelligence. Existing methods utilize patterns, constraints, and machine learning techniques to extract causality, heavily depending on domain knowledge and requiring considerable human effort and time for feature engineering. In this paper, we formulate causality extraction as a sequence labeling problem based on a novel causality tagging scheme. On this basis, we propose a neural causality extractor with the BiLSTM-CRF model as the backbone, named \textbf{SCITE} (\textbf{\textcolor{red}{S}}elf-attentive BiLSTM-\textbf{\textcolor{red}{C}}RF w\textbf{\textcolor{red}{I}}th \textbf{\textcolor{red}{T}}ransferred \textbf{\textcolor{red}{E}}mbeddings), which can directly extract cause and effect without extracting candidate causal pairs and identifying their relations separately. To address the problem of data insufficiency, we transfer contextual string embeddings, also known as Flair embeddings, which are trained on a large corpus in our task. In addition, to improve the performance of causality extraction, we introduce a multihead self-attention mechanism into SCITE to learn the dependencies between causal words. We evaluate our method on a public dataset, and experimental results demonstrate that our method achieves significant and consistent improvement compared to baselines.
\end{abstract}


\end{frontmatter}

\insert\footins{
 \normalfont\footnotesize
  \interlinepenalty\interfootnotelinepenalty
  \splittopskip\footnotesep \splitmaxdepth \dp\strutbox
  \floatingpenalty10000 \hsize\columnwidth
\begin{tabular}{|p{.9\textwidth}|}
 \hline
This is a post-print (accepted manuscript) version of this paper, which has been published in Neurocomputing. This work is licensed under a Creative Commons Attribution-NonCommercial-NoDerivatives 4.0 International License (CC-BY-NC-ND 4.0).
 \begin{center}
 \includegraphics[scale=0.35]{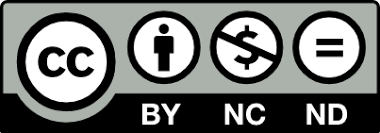}
 \end{center}\\
 \hline
 \end{tabular}}

\section{Introduction}

\begin{figure}
\center
\includegraphics[scale=0.39]{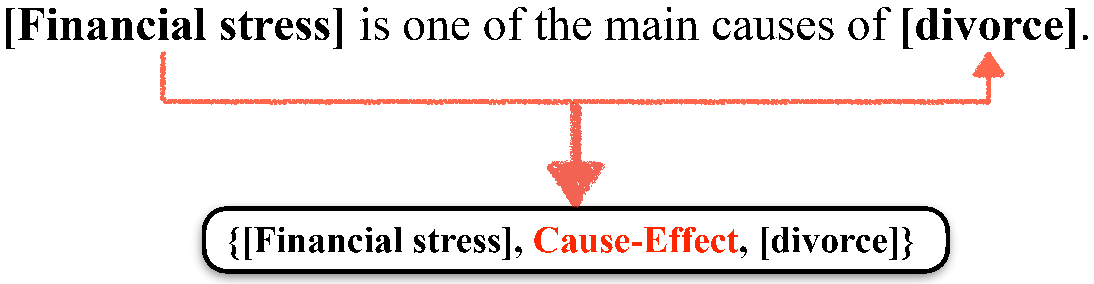}
\caption{A sentence expressing causal relations, in this case, ``financial stress'' is the cause, ``divorce'' is the effect caused by “financial stress”. \label{fig1}}
\end{figure}

Natural language text contains considerably causal knowledge, as shown in Fig.~\ref{fig1}. In recent years, causality extraction has become increasingly
important for many natural language processing tasks, such as information retrieval \cite{khoo1998automatic, Khoo_2001},
event prediction \cite{silverstein2000scalable,radinsky2012learning}, question answering \cite{girju2003automatic,Chang_2006,sobrino2014extracting}, 
generating future scenarios \cite{riaz2010another,hashimoto2014toward}, decision processing \cite{ackerman2012extracting}, 
medical text mining \cite{khoo2000extracting,zhao2018causaltriad,ding2019identification} and behavior 
prediction \cite{Alem_n_Carre_n_2019}. However, due to the ambiguity and diversity
of natural language texts, causality extraction remains a hard NLP problem to solve.

Traditional methods for causality extraction can be divided into
two categories: methods based on patterns \cite{khoo1998automatic, khoo2000extracting, girju2002text,ittoo2011extracting} (Section~\ref{sec5.1}), and methods
based on a combination of patterns and machine learning 
techniques \cite{girju2003automatic,sorgente2013automatic,zhao2016event,luo2016commonsense} (Section~\ref{sec5.2}). The former often has poor cross-domain applicability, fails to
balance precision and recall and may require extensive domain knowledge
to solve problems in a particular area. The latter usually requires considerable
human effort and time on feature engineering, relying heavily on the
manual selection of textual features. Generally, it divides
causality extraction into two subtasks, candidate causal pairs extraction
and relation classification (filtering noncausal pairs). The results
of candidate causal pairs extraction may affect the performance of
relation classification and generate cascading errors.

\citet{zheng2017joint} first proposed a tagging scheme that makes it possible for models to extract entities and relations simultaneously. In their tagging scheme, they apply a cartesian product of the entity mention tags and the relation type tags, and
then assign a unique tag that encodes entity mentions and relation types for each word. Inspired by their novel idea, we focus on a causal triplet that is composed of two event entities and their relation. For instance, the sentence in Fig.~\ref{fig1} contains a causal triplet: ``\{\textbf{financial stress, cause-effect, divorce}\}''. Thus, we can model the causal triplets directly, rather than breaking causality extraction into two subtasks. Based on the motivations, we formulate causality extraction into a sequence tagging problem and propose a causality tagging scheme (Section~\ref{sec2.1}) to achieve direct causality extraction. However, the tagging scheme proposed by \citet{zheng2017joint} cannot identify the overlapping relations in a sentence; it only considers situations where an entity belongs to one triplet: if an entity participates in multiple relations, its tag should not be unique. To address this problem, we design a \textbf{tag2triplet} algorithm (Section~\ref{sec2.2}) to handle multiple causal triplets and embedded causal triplets in the same sentence. Finally, we combine the causality tagging scheme with a deep learning architecture (Section~\ref{sec2.3}) to minimize feature engineering while efficiently modeling causal relations in natural language text.

We notice that some researchers have also proposed deep learning technique-based methods for causality extraction in recent years (Section~\ref{sec5.3}). Although their works are commendable, some works \cite{de2017causal,kruengkrai2017improving,martinez2017neural,Li_2019} are only a classification of causal relations rather than an extraction of complete causal triplets, and others \cite{dasgupta2018automatic,dunietz2018deepcx} mainly focus on the identification of the linguistic expressions for causality instead of the commonsense causality extraction in this paper.

By applying our causality tagging scheme, we use the model based on BiLSTM-CRF \cite{huang2015bidirectional} to extract causal triplets directly. However, we find that two obstacles hinder the further improvement of the performance of the deep learning model.

\textbf{First}, it is difficult to train a superior deep learning model without any prior knowledge in the case of data insufficiency in the existing corpus \cite{hendrickx2009semeval,O_Gorman_2016,mostafazadeh2016caters}. To alleviate this problem, we incorporate Flair
embeddings \cite{akbik2018contextual} into our task, which use the internal states of a character language model trained on a large corpus to create word embeddings (Section~\ref{sec2.3.2}). Experimental results show that this contextual string embedding that has initiated a new technology trend in NLP can drastically improve the performance of causality extraction.

\begin{figure}
\center
\includegraphics[scale=0.35]{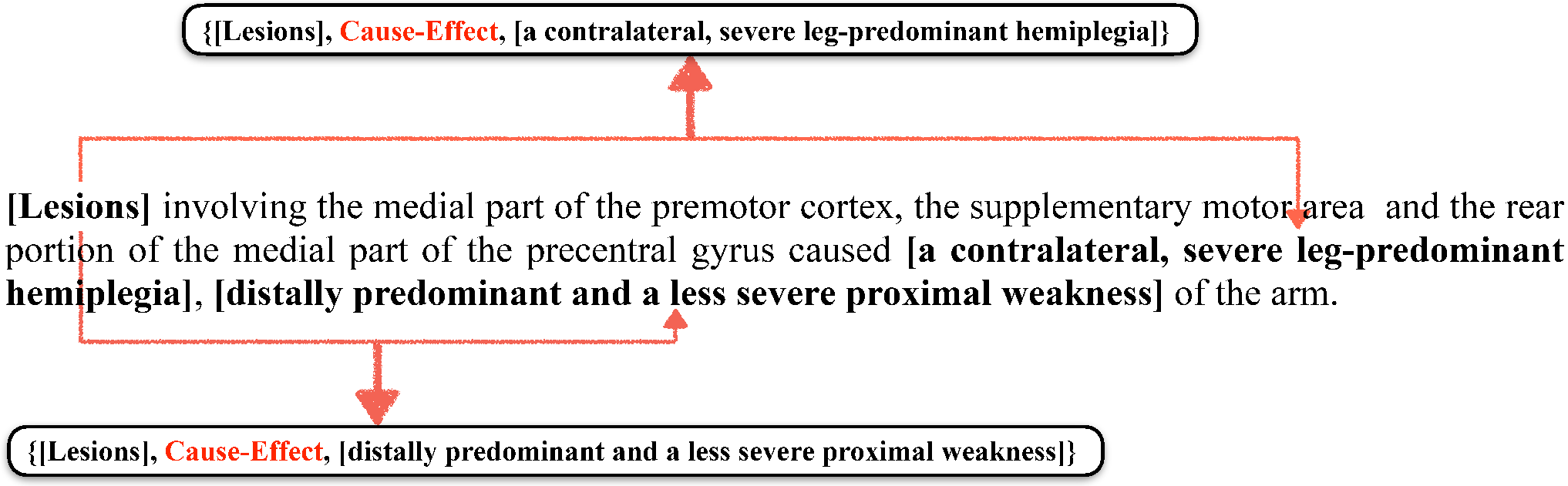}
\caption{The second causal triplet: ``\{\textbf{[lesions], cause-effect, [distally predominant and a less severe proximal weakness]}\}'' spans almost the entire sentence. \label{fig2}}
\end{figure}

\textbf{Second}, in terms of their positions in the text, cause and effect are sometimes far from each other, as Fig.~\ref{fig2} shows. The long-range dependency in the causal triplet creates difficulty and ambiguity in the deep learning model, but a set of logical rules based on dependency trees can easily and accurately extract such triplets. To learn this kind of long-range dependency between cause and effect, we introduce the multihead self-attention mechanism \cite{vaswani2017attention} into our model (Section~\ref{sec2.3.4}). Unlike the LSTM-based model that recursively processes each word, the self-attention mechanism can conduct direct connections between two arbitrary words in a sentence and thus allows unimpeded information flow through the network \cite{tan2018deep}.

The contributions of this paper can be summarized as follows:

\begin{enumerate}
\item We design a novel causality tagging scheme to directly extract causalities in texts and can easily transform the causality extraction into a sequence labeling task and handle multiple causal triplets and embedded causal triplets in the same sentence.
\item Based on our causality tagging scheme, we propose \textbf{SCITE} (\textbf{\textcolor{red}{S}}elf-attentive BiLSTM-\textbf{\textcolor{red}{C}}RF w\textbf{\textcolor{red}{I}}th \textbf{\textcolor{red}{T}}ransferred \textbf{\textcolor{red}{E}}mbeddings), a neural-based causality extractor with transferred contextual string embeddings trained on a large corpus. To the best of our knowledge, we are the first to transfer Flair embeddings into causality extraction.
\item We introduce the multihead self-attention mechanism into SCITE, which enables the model to capture long-range dependencies between cause and effect.
\item Extensive experimental results (Section~\ref{sec3}) and further analysis (Section~\ref{sec4}) show that our method achieves significant and consistent improvement compared to other baselines. We release the code and dataset to the research community for further research \footnote{https://github.com/Das-Boot/scite}.
\end{enumerate}

\section{Method \label{sec2}}

\subsection{Causality Tagging Scheme \label{sec2.1}}

\begin{figure}
\center
\includegraphics[scale=0.35]{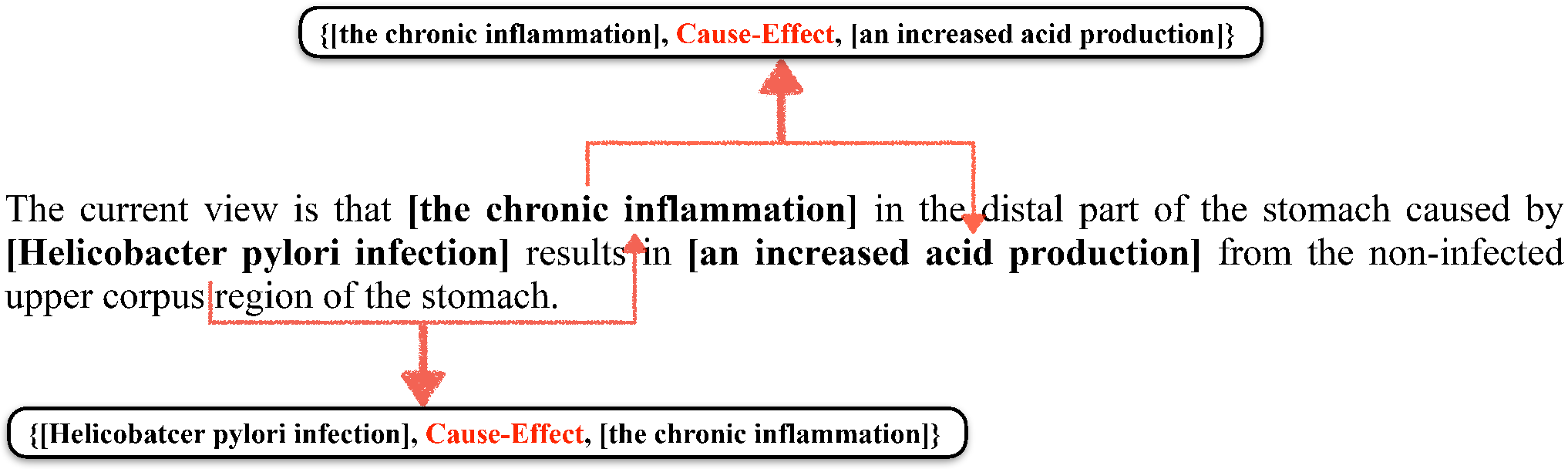}
\caption{Two causal triplets share the same causal event entity: `` \textbf{the chronic inflammation}'' within a sentence.\label{fig3}}
\end{figure}

We use the ``BIO'' (begin, inside, other) and ``C, E, Emb" (cause, effect, embedded causality) signs to represent the position information of the words and the semantic roles of the causal events, respectively, where \textbf{embedded causality} \cite{mostafazadeh2016caters} indicates that a causal event has different roles of causality in different triplets. Fig.~\ref{fig3} is an example of an embedded causality in a sentence. The example sentence contains two causal triplets: ``\{\textbf{the chronic inflammation, cause-effect, an increased acid production}\}'' and ``\{\textbf{Helicobacter, cause-effect, the chronic inflammation}\}'', note that ``\textbf{the chronic inflammation}” is the cause in the first triplet and the effect in the second triplet.

\begin{figure}
\center
\includegraphics[scale=0.34]{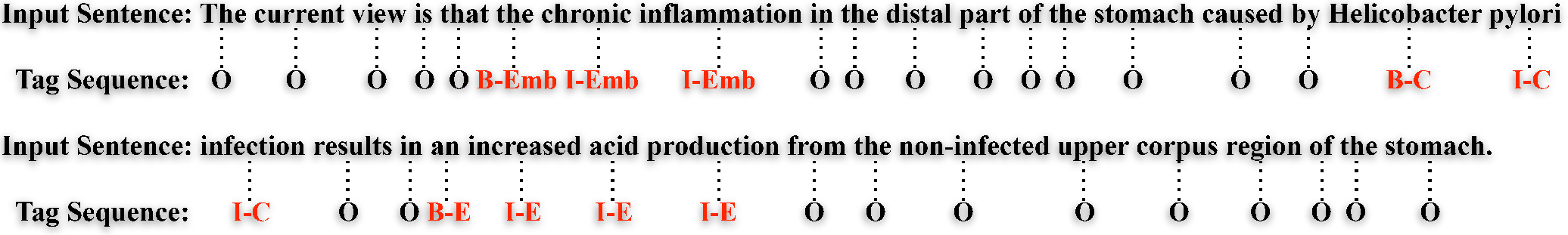}
\caption{A standard annotation for the example sentence based on our causality tagging scheme. \label{fig4}}
\end{figure}

Fig.~\ref{fig4} shows an example of such causality sequence tagging. Based on our causality tagging scheme, we label the causal event entities ``chronic inflammation'', ``Helicobacter pylori infection'' and ``increased acid production'' separately with our special tags. Specifically, tag ``O'' represents the``other'', which means that the corresponding word is irrelevant in any causality components. Tag ``B-C'' represents the ``cause begin'', tag ``I-C'' represents the ``cause inside'', tag ``B-E'' represents the ``effect begin'', tag ``I-E'' represents the ``effect inside'', tag ``B-Emb'' represents the ``embedded causality begin'', and tag ``I-Emb'' represents the ``embedded causality inside''. Thus, the total number of tags is $N_{t}=7$.

\subsection{From Tag Sequence to Causal Triplets \label{sec2.2}}

We design a \textbf{tag2triplet} algorithm for automatically obtaining the final extracted triplets from the tag sequence in Fig.~\ref{fig4}. To better illustrate this algorithm, we define two types of causality: simple causality and complex causality.

\begin{figure}
\center
\includegraphics[scale=0.35]{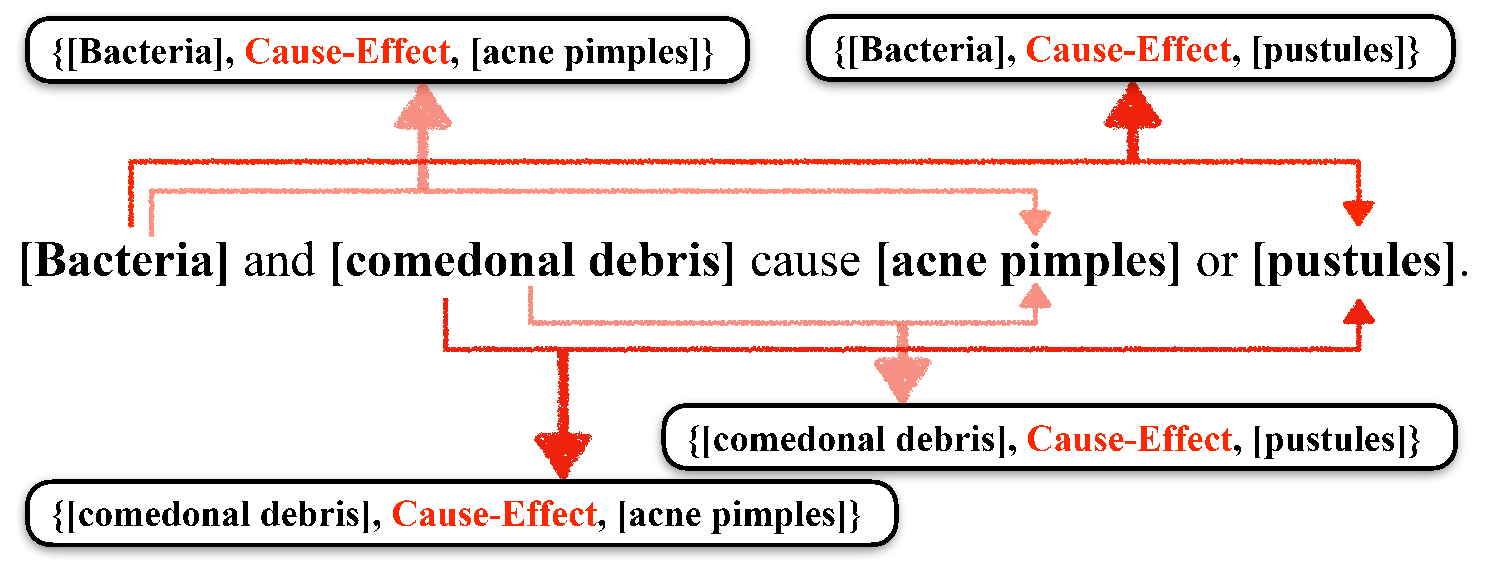}
\caption{For any one of the four causal triplets in the above sentence, there is another causal triplet sharing the same cause or effect. \label{fig5}}
\end{figure}

\subsubsection{The Case of Simple Causality \label{sec2.2.1}}

Simple causality can be classified into two types:
\begin{enumerate}
\item There is only one cause or one effect in the sentence, and there is no embedded causality, that is, $N_{C}=1$ or $N_{E}=1$ and $N_{Emb}=0$, where $N_{C}$, $N_{E}$, and $N_{Emb}$ respectively indicate the number of tags ``B-C'', ``B-E'', and ``B-Emb'' in the sentence. The example sentences in Fig.~\ref{fig1} and Fig.~\ref{fig2} are both of this type of causality.
\item There are multiple causes and effects in the sentence, and there is no embedded causality, i.e., $N_{C}>1$, $N_{E}>1$ and $N_{Emb}=0$. In addition, for each causal triplet in the sentence, there must be at least one causal triplet that shares the same cause or effect. The example sentence in Fig.~\ref{fig5} is this type of causality.
\end{enumerate}

\begin{figure}
\center
\includegraphics[scale=0.35]{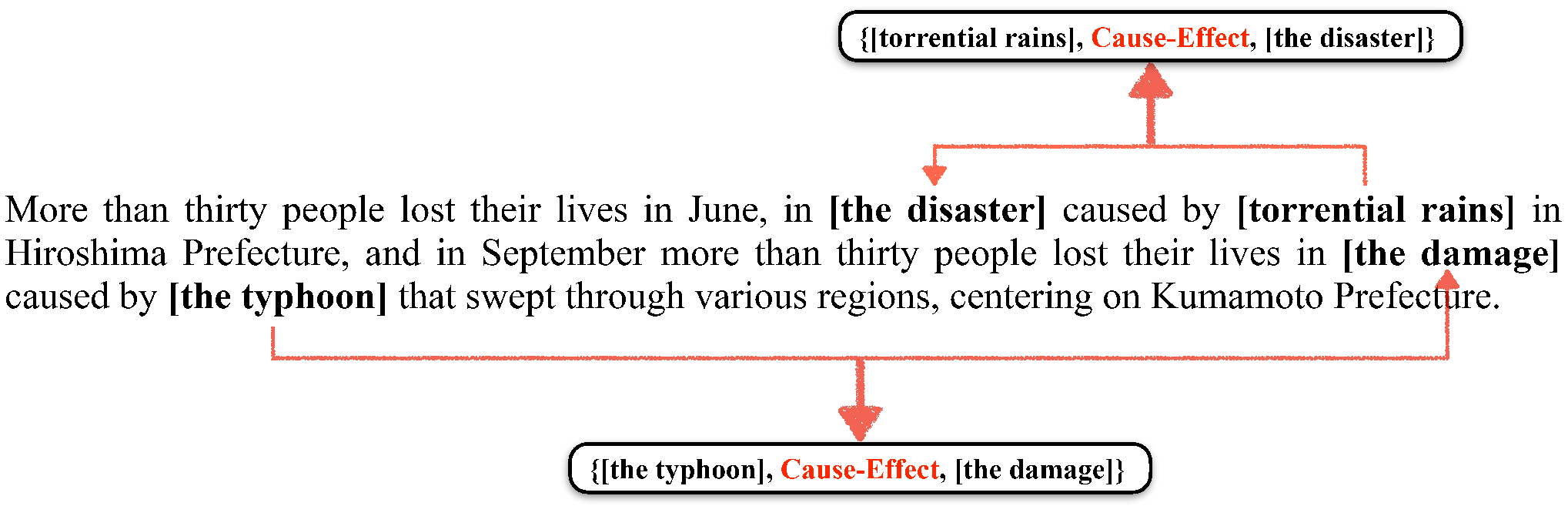}
\caption{The causal triplet ``\{\textbf{torrential rains, cause-effect, the disaster}\}'' and ``\{\textbf{the typhoon, cause-effect, the damage}\}'' do not share the same cause or effect. \label{fig6}}
\end{figure} 

\subsubsection{The Case of Complex Causality \label{sec2.2.2}}

Complex causality has the following two types:
\begin{enumerate}
\item There is embedded causality in the sentence, that is, $N_{C}>0$, $N_{E}>0$ and $N_{Emb}>0$. The example sentence in Fig.~\ref{fig3} is this type of causality.
\item There are multiple causes and effects in the sentence, and there is no embedded causality, i.e., $N_{C}>1$, $N_{E}>1$ and $N_{Emb}=0$. In addition, in all the causal triplets in the sentence, there must be at least one causal triplet that does not share the same cause or effect with any other triplets. The example sentence in Fig.~\ref{fig6} is this type of causality. Note that the distribution of causality in the sentence of Fig.~\ref{fig5} is different from that in Fig.~\ref{fig6}: each causal triplet in the former is mixed together, and each causal triplet in the latter is separated.
\end{enumerate}

\IncMargin{1em}
\begin{algorithm}
\SetKwData{OutDegree}{out-degree}\SetKwData{InDegree}{in-degree}\SetKwData{idx}{idx}\SetKwData{CT}{causal triplets}
\SetKwData{Cand}{candidate}\SetKwData{Can}{candidates}\SetKwData{Flag}{flag}\SetKwData{Dis}{distance}\SetKwData{Rec}{records}
\SetKwFunction{CartesianProduct}{CartesianProduct}\SetKwFunction{Max}{Max}\SetKwFunction{Min}{Min}\SetKwFunction{Sum}{Sum}\SetKwFunction{Len}{Len}\SetKwFunction{App}{AppendToRecord}
\SetKwFunction{Com}{Combination}\SetKwFunction{CD}{CheckDegree}\SetKwFunction{CC}{CheckConjunction}\SetKwFunction{SumDis}{SumDistance}
\SetKwInOut{Input}{input}\SetKwInOut{Output}{output}
\SetKw{Break}{break}
\Input{A tag sequence $S_{tag}$ corresponding to sentence $S$}
\Output{The causal triplets in sentence $S$}
\BlankLine
\emph{Count the \OutDegree and \InDegree for the causality in the $S_{tag}$}\;
\emph{Find the index of causality \idx in the $S_{tag}$}\;
\If{Causality in $S \in $  simple causality}{
 \Cand$\leftarrow$ \CartesianProduct{\idx}\;
 \If{\CC{\Cand, \idx, $S$} is true}{
  \CT$\leftarrow$ \Cand\;
  }
}
\If{Causality in $S \in $ Complex Causality}{
 \Can$\leftarrow$ \CartesianProduct{\idx}\;
 \For{$i\leftarrow \Max{\Sum{\OutDegree}, \Sum{\InDegree}}$ \KwTo $\Len{\Can}$}{
 	\Flag$\leftarrow 0$\;
 	\Rec$\leftarrow [\ ]$\;
 	\For{$j \in \Com{\Can, i}$}{
 	 \If{\CD{$j$, \OutDegree, \InDegree} is true {\bf and} {\CC{$j$, \idx, $S$} is true}}{
 	  \Dis$\leftarrow$ \SumDis{$j$, \idx}\;
 	  \App{$j, \Dis$}\;
 	  \Flag$\leftarrow 1$\;
 	 }
	}
 \If{\Flag $\ne 0$}{ 
  \Break\;
 	 }	
 }
 \CT$\leftarrow$ \Min{\Rec, key=$\Rec[-1]$}\;
}
\caption{Tag2triplet}\label{algo1}
\end{algorithm}\DecMargin{1em}

\begin{table}
\footnotesize
\center
\caption{The intermediate result of running tag2triplet when we input the sentence $S$ and its corresponding tag sequence $S_{tag}$, and we highlight the \textbf{\textcolor{red}{correct}} combination of candidates in bold.}\label{tab1}
\begin{tabular}{@{}cccc@{}}
\toprule
$S$               & \multicolumn{3}{c}{\begin{tabular}[c]{@{}c@{}}... {$[$}the chronic inflammation{$]_{E_{0}}$} ... {$[$}Helicobacter pylori infection{$]_{E_{1}}$} ... \\ {$[$}an increased acid production{$]_{E_{2}}$} ...\end{tabular}} \\
$S_{tag}$ & {$[$}B-Emb I-Emb I-Emb{$]_{E_{0}}$} & {$[$}B-C I-C I-C {$]_{E_{1}}$} & {$[$}B-E I-E I-E I-E{$]_{E_{2}}$} \\ \hline
Index  & {$[5, 6, 7]_{E_{0}}$}                                   & {$[17, 18, 19]_{E_{1}}$}                 & {$[22, 23, 24, 25]_{E_{2}}$}                     \\ 
Out-degree        & $1_{E_{0}}$                                     & $1_{E_{1}}$                 & $0_{E_{2}}$                     \\
In-degree         & $1_{E_{0}}$                                     & $0_{E_{1}}$                 & $1_{E_{2}}$                     \\  
Candidates & ($E_{0}, E_{2}$)                                     & ($E_{1}, E_{0}$)                  & ($E_{1}, E_{2}$)                      \\ \hline
Combinations      & \textbf{\textcolor{red}{($E_{0}, E_{2}$), ($E_{1}, E_{0}$)}} & ($E_{0}, E_{2}$), ($E_{1}, E_{2}$)    & ($E_{1}, E_{0}$), ($E_{1}, E_{2}$)        \\
Out-degree        & \textbf{\textcolor{red}{($1_{E_{0}}$, $1_{E_{1}}$, $0_{E_{2}}$)}}    & ($1_{E_{0}}$, $1_{E_{1}}$, $0_{E_{2}}$)         & ($1_{E_{0}}$, $0_{E_{1}}$, $0_{E_{2}}$)             \\
In-degree         & \textbf{\textcolor{red}{($1_{E_{0}}$, $0_{E_{1}}$, $1_{E_{2}}$)}}     & ($0_{E_{0}}$, $0_{E_{1}}$, $1_{E_{2}}$)         & ($1_{E_{0}}$, $0_{E_{1}}$, $1_{E_{2}}$)             \\ \bottomrule
\end{tabular}
\end{table}

\subsubsection{Tag2triplet Algorithm}

The tag2triplet algorithm is described in Algorithm~\ref{algo1}. We elaborate on the tag2triplet algorithm by taking the sentence $S$ and its corresponding tag sequence $S_{tag}$ as an example in Fig.~\ref{fig4}; the intermediate result is shown in Table~\ref{tab1}. 

First, we count the out-degree and in-degree for causality and find the index of causality in $S_{tag}$. Specifically, the out-degree of ``cause'' is recorded as 1, the in-degree of ``effect'' is recorded as 1, and the out-degree and in-degree of ``embedded causality'' are both recorded as 1. Then, we determine whether the $S$ is simple causality or complex causality according to the number and the distribution of each causal tag: ``C'', ``E'' and ``Emb''. In $S_{tag}$, $N_{Emb} = 1$, and thus, $S$ is complex causality. Then, we apply a \textbf{Cartesian product} of the causal entities composed of causal tags to generate the candidates of the causal triplet. In Table~\ref{tab1}, the candidate ``($E_{0}, E_{2}$)'' represents the causal triplet ``\{\textbf{the chronic inflammation, cause-effect, an increased acid production}\}.

Next, \textbf{combination} returns $i$ length subsequences of triplets from the input candidates. After that, we determine whether the out-degree and in-degree of each combination of candidates are consistent with the original out-degree and in-degree of $S_{tag}$. Then, we determine whether the combination matches the rules according to the coordinating conjunction in $S$, for example, if there is a coordinating conjunction ``and" between adjacent causes in the same clause, then the two causes will form their respective causal triplets with the same effect, as in the ``bacteria" and ``comedonal debris” in Fig.~\ref{fig5}. Finally, we select the combination with the shortest distance from the combinations that passed the checks as the extracted causal triplets. Since only one combination ``($E_{0}, E_{2}$), ($E_{1}, E_{0}$)'' passed all the checks, we directly output it as the final result.

\subsection{SCITE \label{sec2.3}}

\begin{figure}
\center
\includegraphics[scale=0.41]{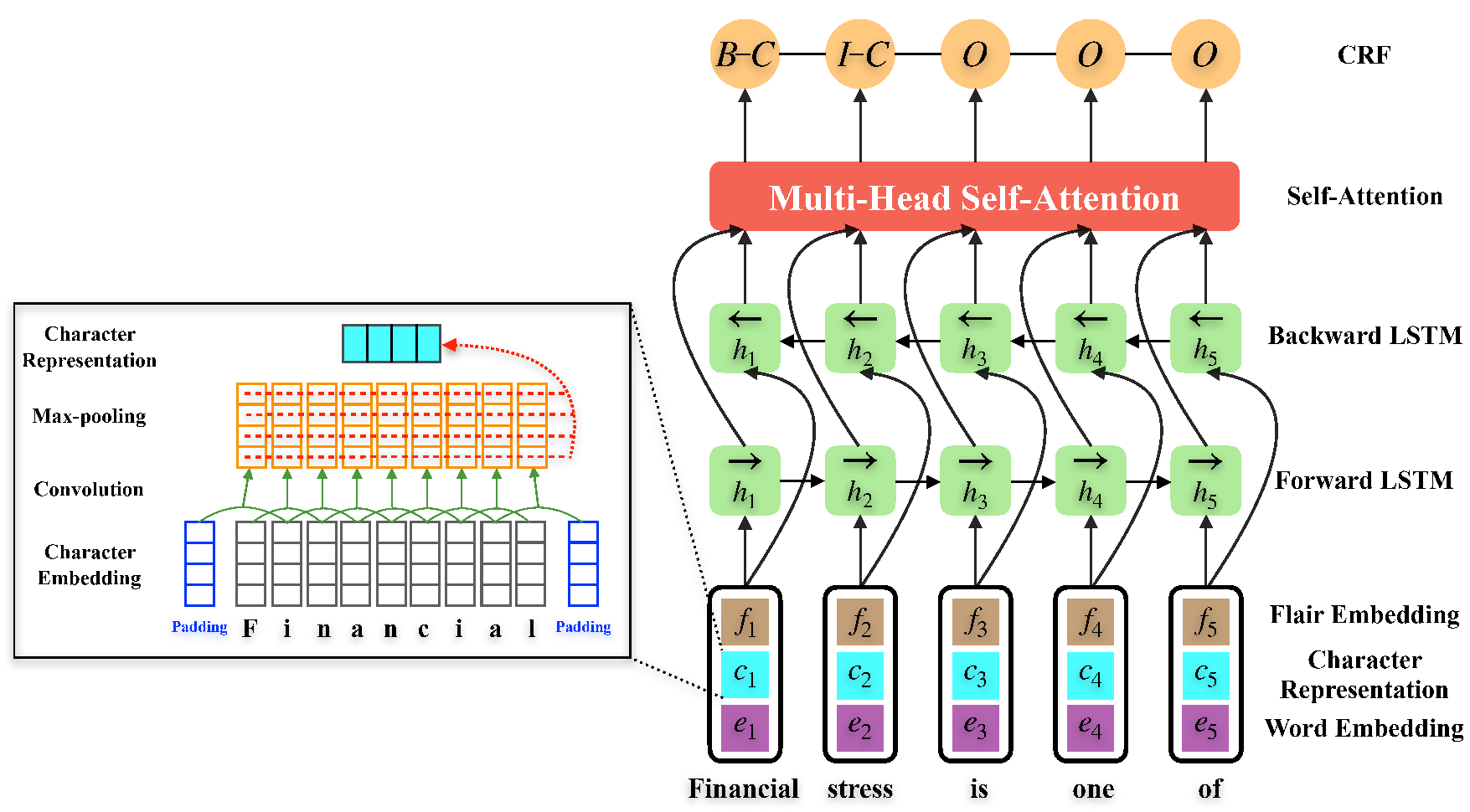}
\caption{The main structure of SCITE for causality sequence labeling. The left side of the figure shows a character CNN structure representing the word ``financial''.  \label{fig7}}
\end{figure}

Fig.~\ref{fig7} gives the main structure of our model SCITE for causality sequence labeling. We take the input sentence $S=\{x_{t}\}^{n}_{t=1}$ and its corresponding label sequence $y=\{y_{i}\}^{n}_{t=1}$ as an example to introduce each component of SCITE from bottom to top as follows, where $n$ is the length of the $S$.

\subsubsection{CNN for Character Representations}

To capture task-specific subword features, we take the same convolutional neural network \cite{lecun1989backpropagation} (CNN) architecture as \citet{ma2016end}, using a one-layer CNN structure followed by a max-over-time pooling operation \cite{collobert2011natural} to learn character-level representations. The process is depicted on the left side of Fig.~\ref{fig7}. 

Specifically, let $\bm{r}_{i} \in \mathbb{R}^{m}$ be the $m$-dimensional character vector corresponding to the $i$-th character in the word $x_{t}$ (the length of $x_{t}$ is $s$). A convolution operation involves a filter $\bm{w} \in \mathbb{R}^{lm}$, which is applied to a window of $l$ characters to produce a new feature. For example, a feature $c_{i}$ is generated from a window of characters $\bm{r}_{i:i+l-1}$ \footnote{In general, let $\bm{r}_{i:i+j}$ refer to the concatenation of character vectors $\bm{r}_{i}$, $\bm{r}_{i+1}$, ..., $\bm{r}_{i+j}$} by

\begin{equation}
c_{i} = \bm{w}^{\mathrm{T}} \bm{r}_{i:i+l-1} + b,
\end{equation}

where $b$ is a bias term. This filter is applied to each possible window of character vectors in the word $\{\bm{r}_{1:l}, \bm{r}_{2:h+1}, ..., \bm{r}_{s-l+1:s}\}$ to produce a feature map

\begin{equation}
\bm{\hat{c}} = [c_{1}, c_{2}, ..., c_{s-l+1}],
\end{equation}

with $\bm{\hat{c}} \in \mathbb{R}^{s-l+1}$. Then, we take the maximum value $\tilde{c}=max\{\bm{\hat{c}}\}$ as the feature corresponding to this particular filter. Thus, denoting that the number of filters is $f$, the character representation $\bm{c}_{t}$ for word $x_{t}$ is given as:

\begin{equation}
\bm{c}_{t} = [\tilde{c_{1}}, \tilde{c_{2}}, ..., \tilde{c_{f}}]
\end{equation}

\subsubsection{Transferring Contextualized Representations Learned from Large Corpus \label{sec2.3.2}}

In recent years, deep learning has ushered in incredible advances in natural language processing (NLP) tasks due to its powerful representation learning ability. However, in the case of data insufficiency of the existing corpus, the data-hungry nature of deep learning limits the performance of our neural-based model in causality extraction. The recent development of contextualized language representation
models \cite{peters2018deep,devlin2018bert,akbik2018contextual} trained on large corpora shed light on the possibility of transfer learning.

In this paper, we use transfer learning to alleviate the problem of data insufficiency. Specifically, we propose to transfer the Flair
embeddings \cite{akbik2018contextual}, which were derived from a character-level language model (CharLM) trained on a 1-billion word benchmark corpus \cite{chelba2013one} to our task. This CharLM consists of a forward language model (fLM) and a backward language model (bLM). Following \citet{akbik2018contextual}, we extract the output hidden state $\overrightarrow{\bm{h}^{t}_{end+1}}$ from the fLM after the last character $r^{t}_{end}$ of the word $x_{t}$. Similarly, we obtain the output hidden state $\overleftarrow{\bm{h}^{t}_{start-1}}$ from the bLM before the first character $r^{t}_{start}$ of the word $x_{t}$. Then, both output hidden states are concatenated to form the final embedding $\bm{f}^{CharLM}_{t}$ of the word $x_{t}$ as follows:

\begin{equation}
\bm{f}^{CharLM}_{t}	=	[\overrightarrow{\bm{h}^{t}_{end+1}}, \overleftarrow{\bm{h}^{t}_{start-1}}]
\end{equation}

Finally, we concatenate transferred Flair embeddings $\bm{f}^{CharLM}_{t}$ and the character representations $\bm{c}_t$ with the word embeddings $\bm{e}_t$ pretrained by \citet{komninos2016dependency} and feed them into a BiLSTM layer.

\subsubsection{BiLSTM}

Long short-term memory (LSTM) \cite{hochreiter1997long} is a particular recurrent neural network (RNN) that overcomes the vanishing and exploding gradient problems \cite{bengio1994learning} of traditional RNN models. Through the specifically designed gate structure of LSTM, the model can selectively save context information.
The basic unit of the LSTM architecture is a memory block, which includes
a memory cell (denoted as $\bm{m}$) and three adaptive multiplication
gates (i.e., an input gate $\bm{i}$, a forget gate $\bm{f}$ and an output
gate $\bm{o}$). Formally, the
computational operations to update an LSTM unit at time $t$ are:

\begin{eqnarray}
\bm{i}_{t}&=&\sigma(\bm{W}_{i}\left[\bm{e}_{t},\bm{c}_{t}, \bm{f}^{CharLM}_{t}\right]+\bm{U}_{i} \bm{h}_{t-1}+\bm{b}_{i}),\\
\bm{f}_{t}&=&\sigma(\bm{W}_{f}\left[\bm{e}_{t},\bm{c}_{t}, \bm{f}^{CharLM}_{t}\right]+\bm{U}_{f} \bm{h}_{t-1}+\bm{b}_{f}), \\
\bm{o}_{t}&=&\sigma(\bm{W}_{o}\left[\bm{e}_{t},\bm{c}_{t}, \bm{f}^{CharLM}_{t}\right]+\bm{U}_{o} \bm{h}_{t-1}+\bm{b}_{o}), \\
\widetilde{\bm{m}_{t}}&=&tanh(\bm{W}_{m}\left[\bm{e}_{t},\bm{m}_{t}, \bm{f}^{CharLM}_{t}\right]+\bm{U}_{m} \bm{h}_{t-1}+\bm{b}_{m}), \\
\bm{m}_{t}&=&\bm{i}_{t}\odot\widetilde{\bm{m}_{t}}+\bm{f}_{t} \odot{\bm{m}_{t-1}}, \\
\bm{h}_{t}&=&\bm{o}_{t}\odot tanh(\bm{m}_{t}), 
\end{eqnarray}

where $[\bm{e}_{t},\bm{c}_{t},\bm{f}^{CharLM}_{t}]$ and $\bm{h}_{t}$ represent the input vector and hidden state, respectively
at time $t$. $\sigma$ is the elementwise sigmoid
function, and $\text{\ensuremath{\odot}}$ is the elementwise product.
$\bm{W}_{i}$, $\bm{W}_{f}$, $\bm{W}_{o}$, $\bm{W}_{m}$ are the weight matrices for the input vector,
$\bm{U}_{i}$, $\bm{U}_{f}$, $\bm{U}_{o}$, $\bm{U}_{m}$ are the weight matrices for the hidden state,
and $\bm{b}_{i}$, $\bm{b}_{f}$, $\bm{b}_{o}$, $\bm{b}_{m}$ denote the bias vectors.

However, LSTM only considers the information
from the past, ignoring future information. To efficiently use contextual
information, we can use bidirectional LSTM (BiLSTM). BiLSTM uses a forward LSTM and a backward LSTM for each
sequence to obtain two separate hidden states: $\overrightarrow{\bm{h}_{t}}$,
$\overleftarrow{\bm{h}_{t}}$, and then the final output at time $t$ is formed by
concatenating these two hidden states:

\begin{equation}
\bm{h_t}=[\overrightarrow{\bm{h}_{t}}, \overleftarrow{\bm{h}_{t}}]
\end{equation}

Therefore, the final output of the BiLSTM layer for the input sentence $S$ can be represented by $\bm{H}=\{\bm{h}_t\}^n_{t=1}$, where $\bm{H} \in \mathbb{R}^{n \times d}$, and $d$ is the layer size of the BiLSTM layer.

\subsubsection{Multihead Self-Attention \label{sec2.3.4}}

\begin{figure}
\center
\includegraphics[scale=0.41]{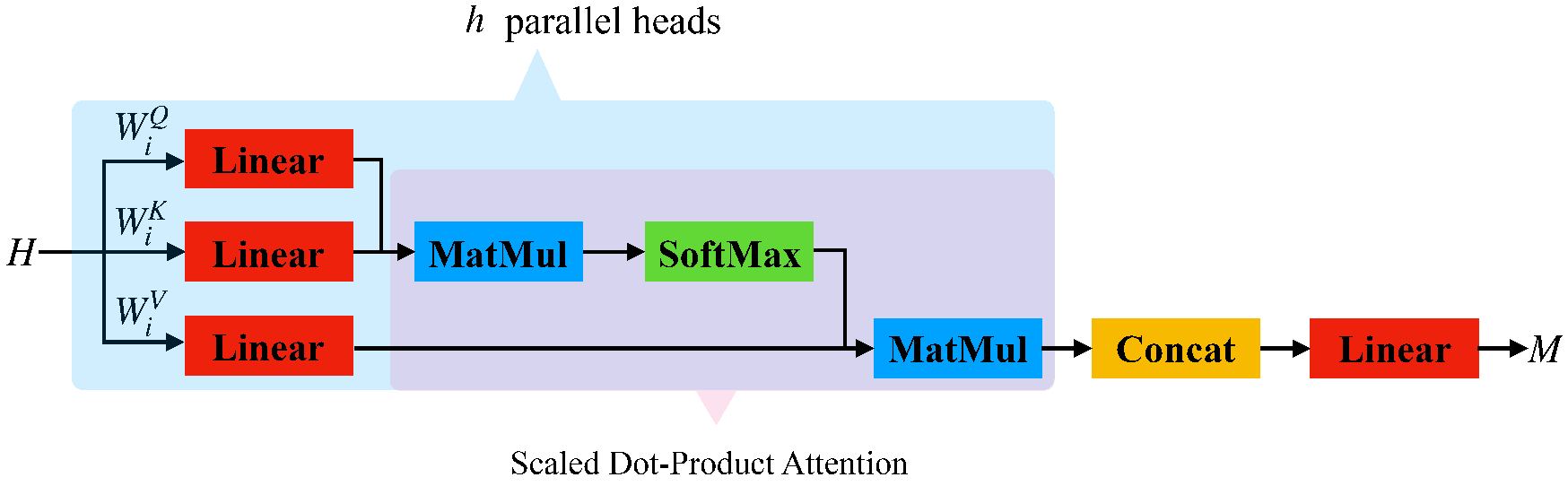}
\caption{The architecture of the multihead attention mechanism.\label{fig8}}
\end{figure}

Self-attention is a particular case of the attention mechanism, which only requires a single sequence to compute its representation, has been successfully applied to many NLP tasks \cite{cheng2016long,lin2017structured,vaswani2017attention} and shows its superiority in capturing long-range dependency. In SCITE, we adopt the multihead self-attention (MHSA) proposed by \citet{vaswani2017attention} to learn the dependencies of causalities in the given sentences. Fig.~\ref{fig8} depicts the architecture of the multihead attention mechanism.

Specifically, given $\bm{H}$ as the output of the BiLSTM layer, the multihead attention mechanism first projects the matrix $\bm{H}$ $h$ times with different learned linear projections to matrices: $\bm{H} \bm{W}_i^Q$, $\bm{H} \bm{W}_i^K$ and $\bm{H} \bm{W}_i^V$. where $h$ is the number of heads and parameter matrices $\bm{W}_i^Q \in \mathbb{R}^{d \times d_v}$, $\bm{W}_i^K \in \mathbb{R}^{d \times d_v}$ and $\bm{W}_i^V \in \mathbb{R}^{d \times d_v}$ are projections for the $i$-th head. Then, the attention function is performed in parallel, yielding $n \times d_v$-dimensional output values. Finally, all the matrices produced by parallel heads are concatenated, resulting in the final values $\bm{M}$ whose dimension is $n \times (hd_{v})$, where both $h$ and $d_v$ are hyperparameters of the self-attention layer. The formulations can be shown as follows:

\begin{align}
\bm{M}=MultiHead(\bm{H}, \bm{H}, \bm{H}) = Concat(\bm{head}_1...,\bm{head}_h) \\
where\ \bm{head}_i = Attention(\bm{H}\bm{W}_i^Q, \bm{H}\bm{W}_i^K, \bm{H}\bm{W}_i^V)
\end{align}

Here, the attention function is the ``scaled dot-product attention'', which computes the attention scores as follows:

\begin{equation}
\resizebox{.9\hsize}{!}{$Attention(\bm{H}\bm{W}_i^Q, \bm{H}\bm{W}_i^K, \bm{H}\bm{W}_i^V)=softmax(\frac{(\bm{H}\bm{W}_i^Q)(\bm{H}\bm{W}_i^K)^\mathrm{T}}{\sqrt{d}})(\bm{H}\bm{W}_i^V)$}
\end{equation}

To fully integrate the information, we concatenate $\bm{H}$ and $\bm{M}$ into matrix $\tilde{\bm{H}}$ and then project $\tilde{\bm{H}}$ with a linear projection to matrix: $\tilde{\bm{H}}\bm{W}$. where weight matrix $\bm{W} \in \mathbb{R}^{(d+hd_{v})k}$ is the parameter of the model to be learned in training and $k$ is the number of distinct tags.

\subsubsection{CRF}

The conditional random field (CRF) \cite{lafferty2001conditional} can obtain a globally optimal chain of labels for a given sequence considering the correlations between adjacent tags. In a sequence labeling task, there are usually strong dependencies between the output labels. Therefore, instead of only using RNN to model tagging decisions separately, we adopt BiLSTM-CRF \cite{huang2015bidirectional} as the backbone of SCIFl to jointly decode labels for the whole sentence.

We use $\bm{P} \in \mathbb{R}^{n \times k}$ as the matrix of scores output by the linear layer, where $\bm{P}_{ij}$ represents the score of the $j^{th}$ label of the $i^{th}$ word within a sentence. For the sentence $S=\{x_t\}_{t=1}^n$ and a path of tags $y=\{y_i\}_{i=1}^n$, CRF gives a real-valued score as follows:

\begin{equation}
score\left(S,y\right)	=	\sum_{i=0}^{n}\bm{A}_{y_{i},y_{i+1}}+\sum_{i=1}^{n}\bm{P}_{i,y_{i}},
\end{equation}

where $\bm{A}$ is the transition matrix, and $\bm{A}_{i,j}$ denotes the score of a transition from tag $i$ to tag $j$. $y_{0}$ and $y_{n}$ are the special tags at the beginning and the end of a sentence, so $\bm{A}$ is a square matrix of size $k+2$. Therefore, the probability for the label sequence $y$ given a sentence $S$ is:

\begin{equation}
p\left(y\mid S\right)	=	\frac{e^{score\left(S,y\right)}}{\sum_{\widetilde{y}\in Y_{S}}e^{score\left(S,\widetilde{y}\right)}},
\end{equation}

We now maximize the log-likelihood of the correct tag sequence:

\begin{equation}
\log\left(p\left(y\mid S\right)\right)	=	score\left(S,y\right)-\log\left(\sum_{\widetilde{y}\in Y_{S}}e^{score\left(S,\widetilde{y}\right)}\right),
\end{equation}

where $Y_S$ represents all possible tag sequences for an input sentence $S$. From the formulation above, we can obtain a valid output sequence. When decoding, the sequence with the maximum score is output by:

\begin{equation}
y^{*} = arg\max_{\widetilde{y}\in Y_{S}}score\left(S,\widetilde{y}\right)
\end{equation}

This can be computed using dynamic programming techniques, and we choose the Viterbi algorithm \cite{DBLP:journals/tit/Viterbi67} for this decoding.

\section{Experiments\label{sec3}}

\subsection{Experimental Settings}

\subsubsection{Dataset}

In the experiment, we evaluate a corpus obtained by extending the annotations of the SemEval 2010 task 8 dataset.
\cite{hendrickx2009semeval}. In the original dataset, only one causal triplet in each sentence was annotated. We extend the annotation with the causal triplets not considered by the SemEval annotators; for example, we annotate all of the causal triplets in the sentence in Fig.~\ref{fig2} (more examples are shown in Fig.~\ref{fig3}, Fig.~\ref{fig5} and Fig.~\ref{fig6}). Specifically, the corpus is composed of 5,236 sentences, of which 1,270 sentences contain at least one causal triplet. The training set consists of 4,450 sentences and contains 1,570 causal triplets. There are 804 sentences in the test set, including 296 causal triplets. Table~\ref{tab2} shows the statistics of six types of causal tags for the dataset.

\begin{table}
\small                                                                 
\center
\caption{Statistics of different types of causal tags for the dataset}\label{tab2}
\begin{tabular}{ccc}
\toprule
Tag Type & Training Set                & Test Set \\ \hline
B-C      & 1308                        & 236      \\
I-C      & 1421                        & 229      \\
B-E      & 1268                        & 238      \\
I-E      & 1230                        & 230      \\
B-Emb    & {\color[HTML]{333333} 55}   & 9        \\
I-Emb    & {\color[HTML]{333333} 55}   & 16       \\ \hline
Sum      & {\color[HTML]{333333} 5337} & 958      \\ \bottomrule
\end{tabular}
\end{table}

\subsubsection{Evaluation}

We use standard precision (P), recall (R) and F1-score (F) as evaluation metrics, which can be calculated by the following formulas:

\begin{eqnarray}
P&=&\frac{\text{\#correct extracted causal triplets}}{\text{\#extracted causal triplets}},\\
R&=&\frac{\text{\#correct extracted causal triplets}}{\text{\#total causal triplets in D}}, \\
F&=&2\frac{P\cdot R}{P+R}, 
\end{eqnarray}

where $D$ is the set of all the sentences in the dataset and a predicted causal triplet is regarded as correct if and only if it precisely matches a labeled causal triplet. To obtain comparable and reproducible F1-scores, we follow the advice of \citet{reimers2017reporting} and conduct each experiment 5 times and then report the average results and their standard deviation, as shown in Table~\ref{tab3}.

\subsubsection{Hyperparameters}

The model is implemented by using Keras \footnote{https://github.com/keras-team/keras} version 2.2.4. The 300-D word embeddings pretrained by \citet{komninos2016dependency} are employed and kept fixed during the training process. Character embeddings are randomly initialized from a uniform distribution ranging in $[-\sqrt{\frac{3}{dim}}, +\sqrt{\frac{3}{dim}}]$, where we set $dim=30$. For the character-level CNN layer, we use a one-layer CNN with 30 filters, and the window size is 3. We use the Flair framework \footnote{https://github.com/zalandoresearch/flair} to compute the Flair embeddings. The hidden size of LSTM is set to 256. The parameters $h$ (the number of heads) and $d_{v}$ (the size of each head) of the multihead self-attention mechanism are set to 3 and 8, respectively. We use variational dropout \cite{gal2016theoretically} with a dropout rate of 0.5 to regularize our network. To address the exploding gradient problem, we
apply gradient normalization \cite{pascanu2013difficulty} with a threshold of 5.0 to the SCITE. The optimization method of the training process is Nadam \cite{dozat2016incorporating} with a learning rate of 0.001, and we apply a learning rate annealing method such that if the training loss does not fall for more than 10 epochs, this method will halve the learning rate. We let the minibatch size be 16. In the experiments, we perform a grid search and 10-fold cross-validation on the training set to find the optimal hyperparameters. On the test set, we select the optimal model among all 200 epochs with the highest cross-validation F1-score.

\subsubsection{Baselines}

For a comprehensive comparison, we compare our method against several classic causality extraction methods, which can be divided into two categories: pipeline methods and sequence tagging models based on our causality tagging scheme. The pipeline methods that we use as our baselines are as follows:
\begin{itemize}
\item \textbf{Rules+Bayesian}: \citet{sorgente2013automatic} performed pattern matching to extract candidate cause-effect pairs based on a set of rules and then used a Bayesian classifier and Laplace smoothing to filter noncausal pairs.
\item \textbf{CausalNet}: \citet{luo2016commonsense} proposed causal strength (CS) to measure the causal strength between any two pieces of short texts, integrating necessity causality with sufficiency causality. For comparison, we add the same cause-effect extraction module
as \citet{sorgente2013automatic} to their method. We then calculate the CS score of the candidate causal pair and compare it with the threshold $\tau$ ($\tau$ is a tunable hyperparameter). If $CS\left(c,e\right)>\tau$, we conclude that $(c,e)$ is a causal relation; otherwise, $(c,e)$ is an erroneously extracted pair.
\end{itemize}

The sequence tagging structure used in this paper is divided into CNN-based models and BiLSTM-based models. For the CNN-based models \cite{Strubell2017Fast}, the baselines are as follows:

\begin{itemize}
\item \textbf{IDCNN-Softmax}: This model uses a deep iterated dilated CNN (IDCNN) architecture to aggregate context from the entire text, which has better capacity than traditional CNN and faster computational speed than LSTM, and then map the output of IDCNN to predict each label independently through a softmax classifier.
\item \textbf{IDCNN-CRF}: This model uses the CRF classifier to maximize the label probability of the complete sentence based on IDCNN. Compared to
the softmax classifier, the CRF classifier is more appropriate for tasks with strong output label dependency.
\end{itemize}

The baselines for the BiLSTM-based models are listed as follows:

\begin{itemize}
\item \textbf{BiLSTM-softmax} \cite{wang2015part}: The model consists of two parts: a BiLSTM encoder and a softmax classifier.
\item \textbf{BiLSTM-CRF} \cite{huang2015bidirectional}: A classic and popular choice for sequence labeling tasks, which consists of a BiLSTM encoder and a CRF classifier.
\item \textbf{CLSTM-BiLSTM-CRF} \cite{lample2016neural}: A hierarchical BiLSTM-CRF model that uses character-based representations to implicitly capture morphological features (e.g., prefixes and suffixes) through a character LSTM encoder (CLSTM) and then concatenates the character embeddings and pretrained word embeddings as the input of BiLSTM-CRF.
\item \textbf{CCNN-BiLSTM-CRF} \cite{ma2016end}: A similar hierarchical BiLSTM-CRF model uses a character CNN encoder (CCNN) instead of a CLSTM to learn the character-level embeddings.
\end{itemize}

To further analyze the performance of Flair embeddings transferred into our task, we combine the ELMo \cite{peters2018deep} and BERT \cite{devlin2018bert}, two powerful contextualized word representations, into our task-specific BiLSTM-CRF architecture as the experimental baselines:

\begin{itemize}
\item \textbf{ELMo-BiLSTM-CRF}: An extension of BiLSTM-CRF in which \citet{peters2018deep} concatenate pretrained static word embeddings with the ELMo (Embeddings from Language Models) representations and take them as the input of BiLSTM-CRF.
\item \textbf{BERT-BiLSTM-CRF}: A similar extension in which \citet{devlin2018bert} added pretrained word embeddings and the BERT (bidirectional encoder representations from transformers) representations and used them as the input of BiLSTM-CRF.
\item \textbf{Flair-BiLSTM-CRF}: This model is used as a strong baseline in our work, in which \citet{akbik2018contextual} pretrained word embeddings are concatenated with the Flair embeddings and fed it into the BiLSTM-CRF model. Note that the models using Flair embeddings have achieved the current state-of-the-art results in a range of sequence labeling tasks such as named entity recognition, chunking and part-of-speech tagging \cite{akbik2018contextual,Borchmann2018}.
\item \textbf{Flair+CLSTM-BiLSTM-CRF}: A simple extension in which \citet{akbik2018contextual} added task-trained character representations learned from a CLSTM to Flair-BiLSTM-CRF.
\end{itemize}

\subsection{Experimental Results}

The performance of different models on the causality extraction is shown in Table~\ref{tab3}. The first part is the pipeline methods (from row 2 to row 3). The second part (row 4 to row 5) is the CNN-based sequence tagging method. The third part (row 6 to row 9) is the BiLSTM-based sequence tagging method, and the fourth part (row 10 to row 13) is the sequence tagging method using contextualized word embeddings. Our SCITE model is shown in the final part, where the first row is the result of SCITE based on the proposed tagging scheme and the second row is the result of SCITE based on the general tagging scheme.

Table~\ref{tab3} shows that SCITE outperforms all other models with an F1-score of 0.8455 in the test set. This demonstrates the effectiveness of our proposed method. Furthermore, it also shows that the sequence tagging models are better than pipeline methods.

\begin{table}
\setlength{\tabcolsep}{2.48pt}
\footnotesize
\center
\caption{Comparison in precision (P), recall (R), and F1-score (F) on the test set with
baselines.}\label{tab3}
\begin{tabular}{cccc}
\toprule
Model                  & P                                             & R                                 & F                                 \\ \hline
CausalNet              & 0.6211                                        & 0.5372                            & 0.5761                            \\
Rules-Bayesian         & 0.6042                                        & 0.5878                            & 0.5959                            \\ \hline
IDCNN-softmax          & 0.7455$\pm$0.0142                             & 0.7074$\pm$0.0168                     & 0.7258$\pm$0.0105                    \\
IDCNN-CRF              & 0.7442$\pm$0.0225                                & 0.7142$\pm$0.0122                     & 0.7288$\pm$0.0160                     \\ \hline
BiLSTM-softmax         & 0.7744$\pm$0.0183                                 & 0.7622$\pm$0.0114                     & 0.7682$\pm$0.0138                     \\
CLSTM-BiLSTM-CRF       & 0.8144$\pm$0.0284             & 0.7412$\pm$0.0073                          & 0.7757$\pm$0.0107 \\
CCNN-BiLSTM-CRF        & 0.8069$\pm$0.0199                                 & 0.7520$\pm$0.0227                     & 0.7780$\pm$0.0075                     \\
BiLSTM-CRF             & 0.7837$\pm$0.0061                                 & 0.7932$\pm$0.0087                     & 0.7884$\pm$0.0072                     \\
 \hline
BERT-BiLSTM-CRF        & 0.8277$\pm$0.0058                                 & 0.8209$\pm$0.0093                     & 0.8243$\pm$0.0049                     \\
Flair+CLSTM-BiLSTM-CRF & 0.8403$\pm$0.0090          & 0.8284$\pm$0.0125                     & 0.8343$\pm$0.0106                     \\
ELMo-BiLSTM-CRF        & 0.8361$\pm$0.0135                                 & 0.8399$\pm$0.0063                     & 0.8379$\pm$0.0092                     \\
Flair-BiLSTM-CRF       & \textbf{0.8414$\pm$0.0079}          & 0.8351$\pm$0.0141                     & 0.8382$\pm$0.0092                     \\ \hline
\multirow{2}{*}{\begin{tabular}[c]{@{}c@{}}\textbf{SCITE}\\ (Flair+CCNN-BiLSTM-MHSA-CRF)\end{tabular}} & \multirow{2}{*}{0.8333$\pm$0.0042} & \multirow{2}{*}{\textbf{0.8581$\pm$0.0021}           } & \multirow{2}{*}{\textbf{0.8455$\pm$0.0028}} \\
                                                               &                    &                    &                    \\ 
                                                               \multirow{2}{*}{\begin{tabular}[c]{@{}c@{}}SCITE\\ (based on general tagging scheme)\end{tabular}} & \multirow{2}{*}{0.7609$\pm$0.0170} & \multirow{2}{*}{0.7757$\pm$0.0136}            & \multirow{2}{*}{0.7682$\pm$0.0145} \\
                                                               &                    &                    &                    \\ 
\bottomrule
\end{tabular}
\end{table}

By comparing the performance of the sequence tagging models on the test set, we can see that the BiLSTM-based models are better than the CNN-based models. The reason for the superior performance of BiLSTM-based models may be that the LSTM layer can more efficiently capture the global word context information and learn semantic representations of causality. In addition,
It also shows that the performance of the model drastically improves after feeding contextualized word representations into the BiLSTM-CRF architecture. In particular, the Flair-BiLSTM-CRF achieves the highest improvement of 6.32\% over the BiLSTM-CRF compared with the ELMo and BERT (increases of 6.28\% and 4.55\%, respectively), which indicates that the contextualized character-level word embedding is more suitable for the task of causality extraction.

Moreover, we also find that our proposed causality tagging scheme yields a better result than the general tagging scheme (0.8455 versus 0.7682) under the SCITE architecture, which verifies the effectiveness of our proposed tagging scheme. The general tagging scheme does not contain an ``Emb'' tag, and thus, the model cannot correctly identify embedded causality. Although the number of embedded causalities is relatively small in the test set, embedded causality plays a crucial role in causality extraction: one error in its identification may affect the correct extraction of multiple triplets, as shown in Fig.~\ref{fig3}.

\section{Analysis and Discussion \label{sec4}}

\subsection{Error Analysis}

\begin{figure}
\center
\includegraphics[scale=0.44]{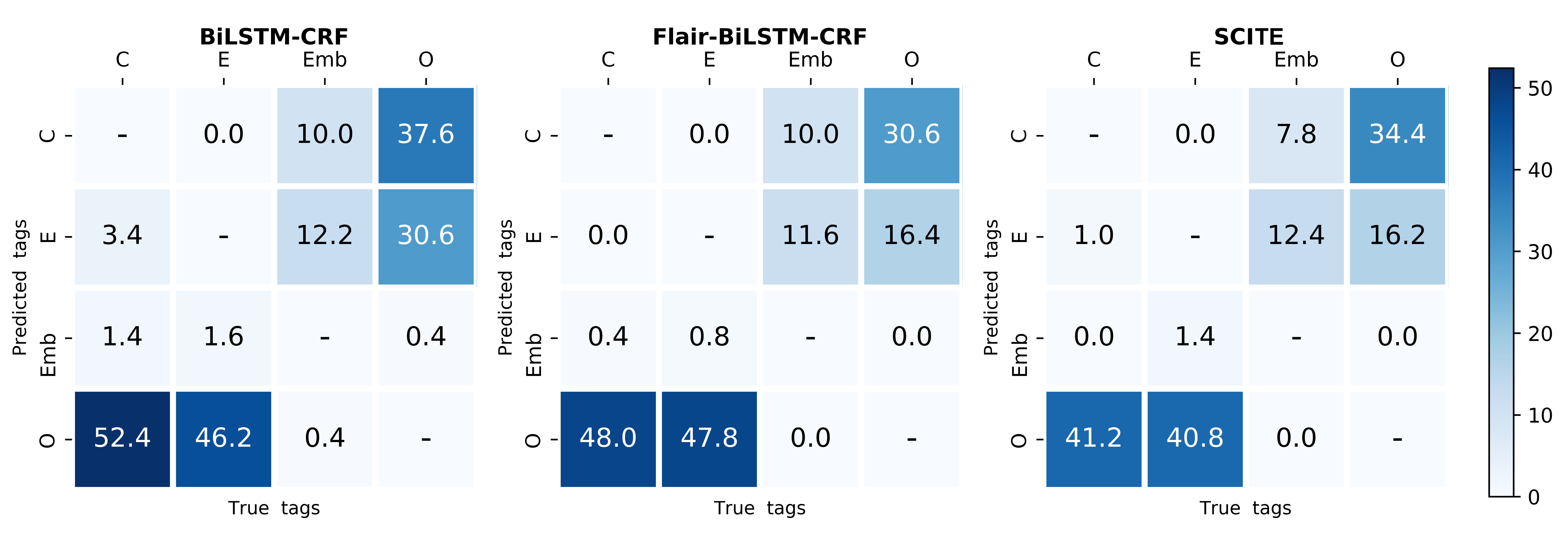}
\caption{Confusion matrix of our SCITE model and other baseline models for tag errors. x-axis: true tags; y-axis: predicted tags.\label{fig9}}
\end{figure}

In this paper, we focus on extracting all causal triplets from natural language texts, where the accurate identification of tags ``C'' (cause), ``E'' (effect) and ``Emb'' (embedded causality), which represent the semantic roles of causal events, plays a vital role in our task. To perform error analysis, we present a confusion matrix for tags ``C'' (including ``B-C'' and ``I-C''), ``E'' (including ``B-E'' and ``I-E''), and ``Emb'' (including ``B-Emb'' and ``I-Emb'') shown in Fig.~\ref{fig9}. We can see that most of the errors are confusion between ``C'', ``E'' and ``O''. This confusion may arise due to the problem of insufficient annotated data. Compared with other baselines \footnote{For the convenience of the display, we only show the results of SICFI, Flair-BiLSTM-CRF (the superior of baselines) and BiLSTM-CRF (the classic sequence tagging model). \label{ft5}}, our model SCITE can better identify ``C'', ``E'', and ``Emb''.

Furthermore, we compare the tagwise performance of our SCITE model with baselines. The comparative results are summarized in Table~\ref{tab4}. First, we observe that our model achieves No. 1 in tags ``C'' (including ``B-C'' and ``I-C''), ``E'' (including ``B-E'' and ``I-E''), and ``Emb'' (including ``B-Emb'' and ``I-Emb'') in terms of F1-score. Second, we also notice that the F1-scores are approximately 0.9 except for tag ``Emb'' because of its low frequency (only 110 instances) in the training set. In particular, it can be seen from the confusion matrix in Fig.~\ref{fig9} that most ``Emb'' tags in the test set are misidentified as ``C'' or ``E'', which leads to the low recall of ``Emb''.

\begin{table}
\setlength{\tabcolsep}{2.48pt}
\scriptsize
\center
\caption{Comparison of predicted tags concerning ``C'' (cause), ``E'' (effect) and ``Emb'' (embedded causality) in precision (P), recall (R), and F1-score (F) on the test set.}
\label{tab4}
\begin{tabular}{cccccccccc}
\toprule
Model            & C-P    & C-R    & C-F    & E-P    & E-R    & E-F    & Emb-P  & Emb-R  & Emb-F  \\ \hline
BiLSTM-CRF       & 0.8810 & 0.8628 & 0.8718 & 0.8928 & 0.8897 & 0.8913 & 0.4343 & 0.0960 & 0.1567 \\
Flair-BiLSTM-CRF & 0.8995 & 0.8843 & 0.8917 & \textbf{0.9294} & 0.8885 & 0.9084 & \textbf{0.8556} & 0.1360 & 0.2197 \\
\textbf{SCITE}            & \textbf{0.8999} & \textbf{0.8998} & \textbf{0.8998} & 0.9272 & \textbf{0.9021} & \textbf{0.9144} & 0.8489 & \textbf{0.1920} & \textbf{0.2947} \\ 
\bottomrule
\end{tabular}
\end{table}

\subsection{Ablation Analysis}

To investigate the effect of the different components in SCITE (Flair+CCNN-BiLSTM-MHSA-CRF), we also report the results of ablation experiments in Table~\ref{tab5}. All parts positively contribute to the performance of the SCITE model.

\begin{table}
\footnotesize
\center
\caption{Ablation analysis of our proposed model SCITE. ``All'' denotes the complete SCITE model, i.e., the Flair+CCNN-BiLSTM-MHSA-CRF model, while ``-'' denotes removing the component from the SCITE. \label{tab5}}
\begin{tabular}{ccc}
\toprule
Model        & Setting                       & F                          \\ \hline
\textbf{SCITE}  & \textbf{All} & \textbf{0.8455} \\
Flair-BiLSTM-MHSA-CRF  & -CCNN     & 0.8438           \\
Flair-BiLSTM-CRF   & -CCNN -MHSA           & 0.8382\\
BiLSTM-MHSA-CRF  & -Flair -CCNN           & 0.8137         \\
BiLSTM-CRF   & -Flair -CCNN -MHSA                  & 0.7884         \\ 
\bottomrule
\end{tabular}
\end{table}

\begin{figure}
\center
\includegraphics[scale=0.5]{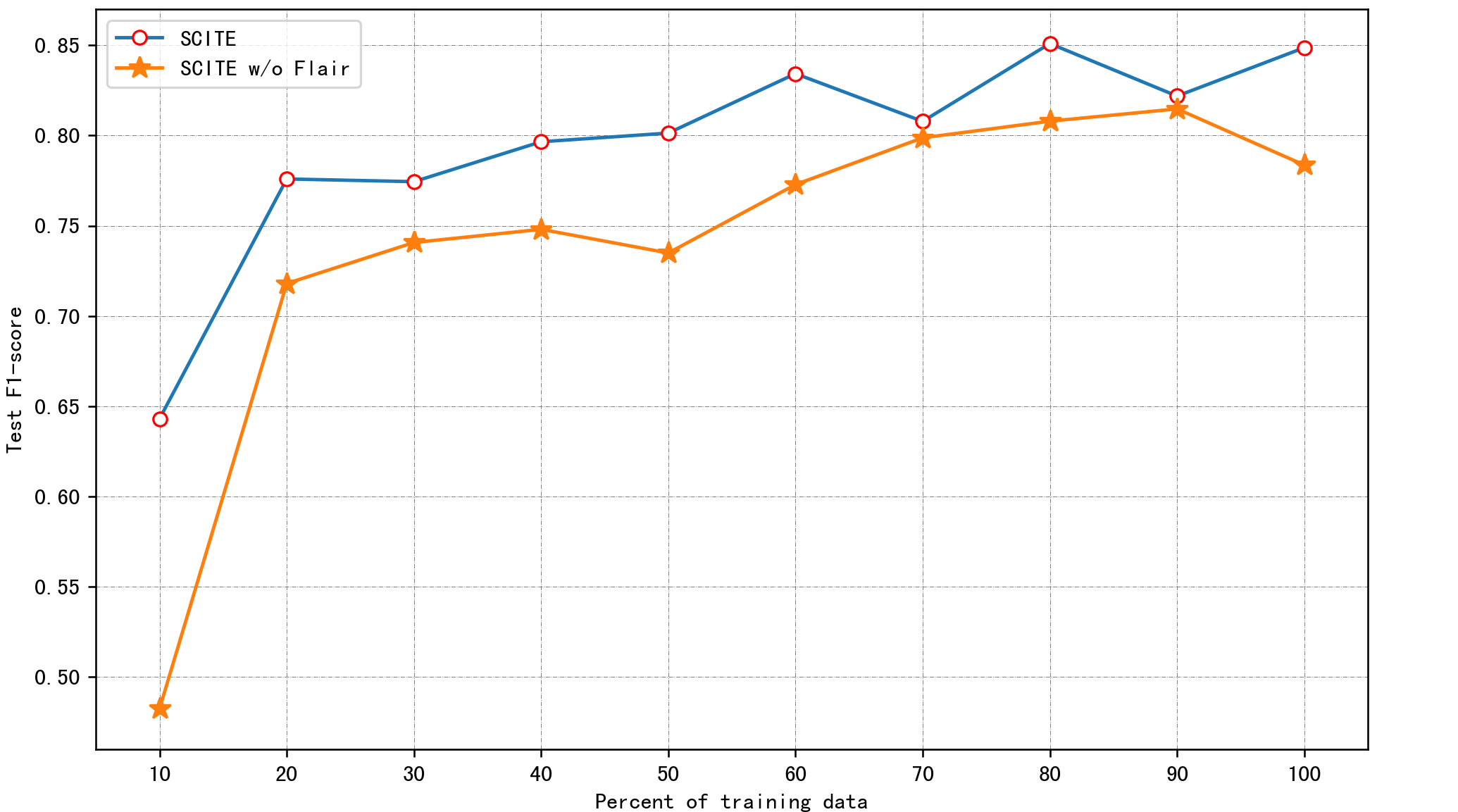}
\caption{F1-score on the test set, in terms of the size of the training dataset. \label{fig10}}
\end{figure}

Specifically, we find that the transferred Flair embeddings provide the most significant improvement. This validates our assumption that the lack of data containing causal triplets in the existing corpus will affect the performance of a neural-based model in causality extraction. Impressively, compared with SCITE without Flair embeddings (SCITE-Flair), the transferred Flair embeddings achieve an improvement of 33.28\% in terms of the F1-score in the case of extremely annotated data insufficiency (10\% of training data), as shown in Fig.~\ref{fig10}. With the help of the transferred contextualized representations, we can not only learn more semantic and syntactic information from the text but also capture word meaning in context to address the polysemous and context-dependent nature of words.

In addition, we also find that the multihead self-attention (MHSA) mechanism can further improve performance, especially when there are no Flair embeddings; the reason is discussed in Section~\ref{sec4.2}. Finally, we find that the task-specific character features can also influence the performance of the model by a slight increase when comparing the models with and without the character representations learned from a CCNN.

\subsection{Analysis of Multihead Self-Attention\label{sec4.2}}

Different from other sequence tagging models, SCITE uses the multihead self-attention mechanism to learn the dependencies between cause and effect. To further analyze the effect of the MHSA, we compute and visualize the F1-score in terms of causality distance (the distance between cause and effect) for the three groups of models:

\begin{itemize}
\item Group 1: BiLSTM-CRF and BiLSTM-MHSA-CRF;
\item Group 2: Flair-BiLSTM-CRF and Flair-BiLSTM-MHSA-CRF;
\item Group 3: Flair+CCNN-BiLSTM-CRF and SCITE (Flair+CCNN-BiLSTM-MHSA-CRF)
\end{itemize}

\begin{figure}
	\center
    \subfloat[Group 1]{\label{fig11a}\includegraphics[scale=0.365]{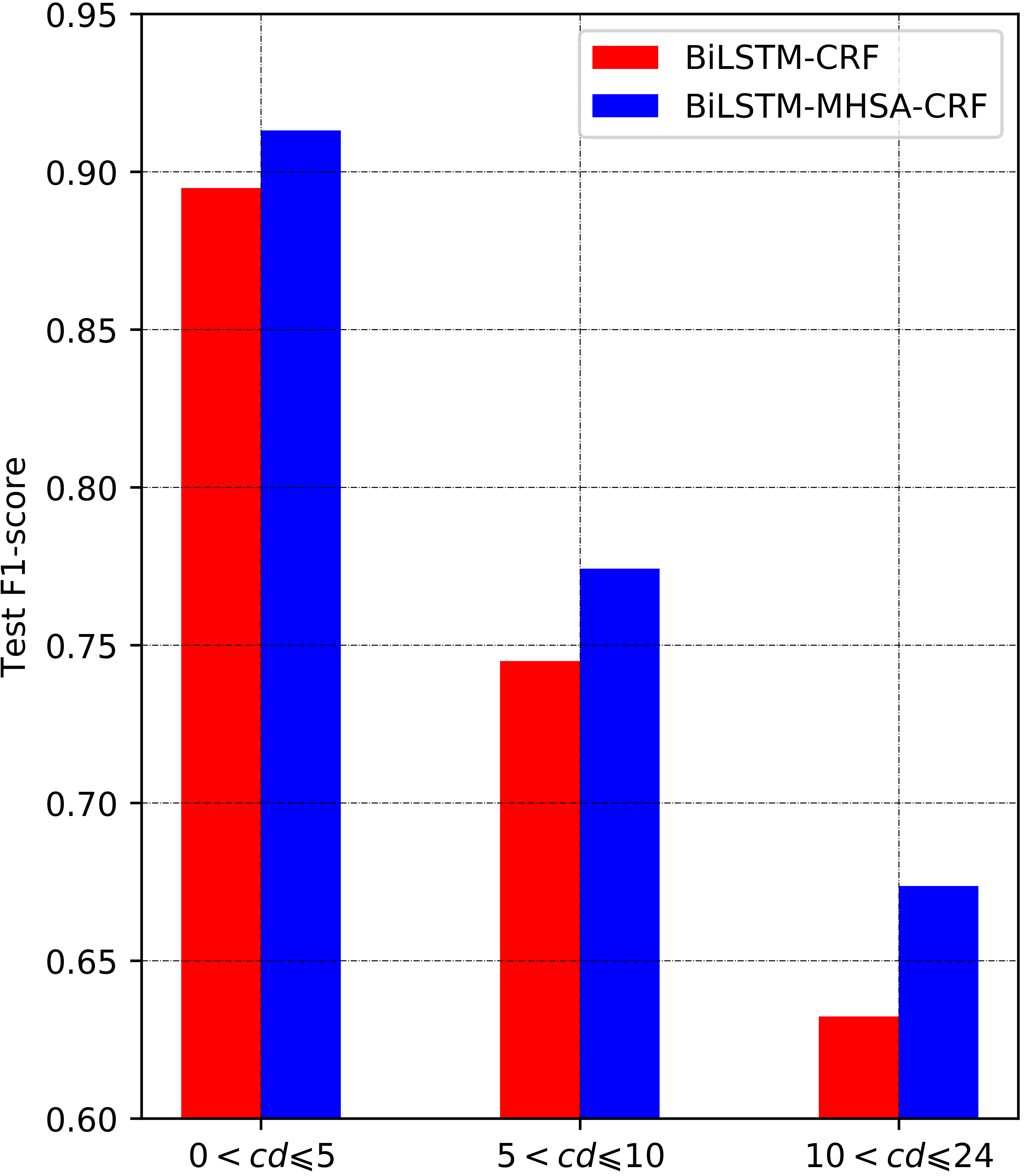}} 
    \subfloat[Group 2]{\label{fig11b}\includegraphics[scale=0.365]{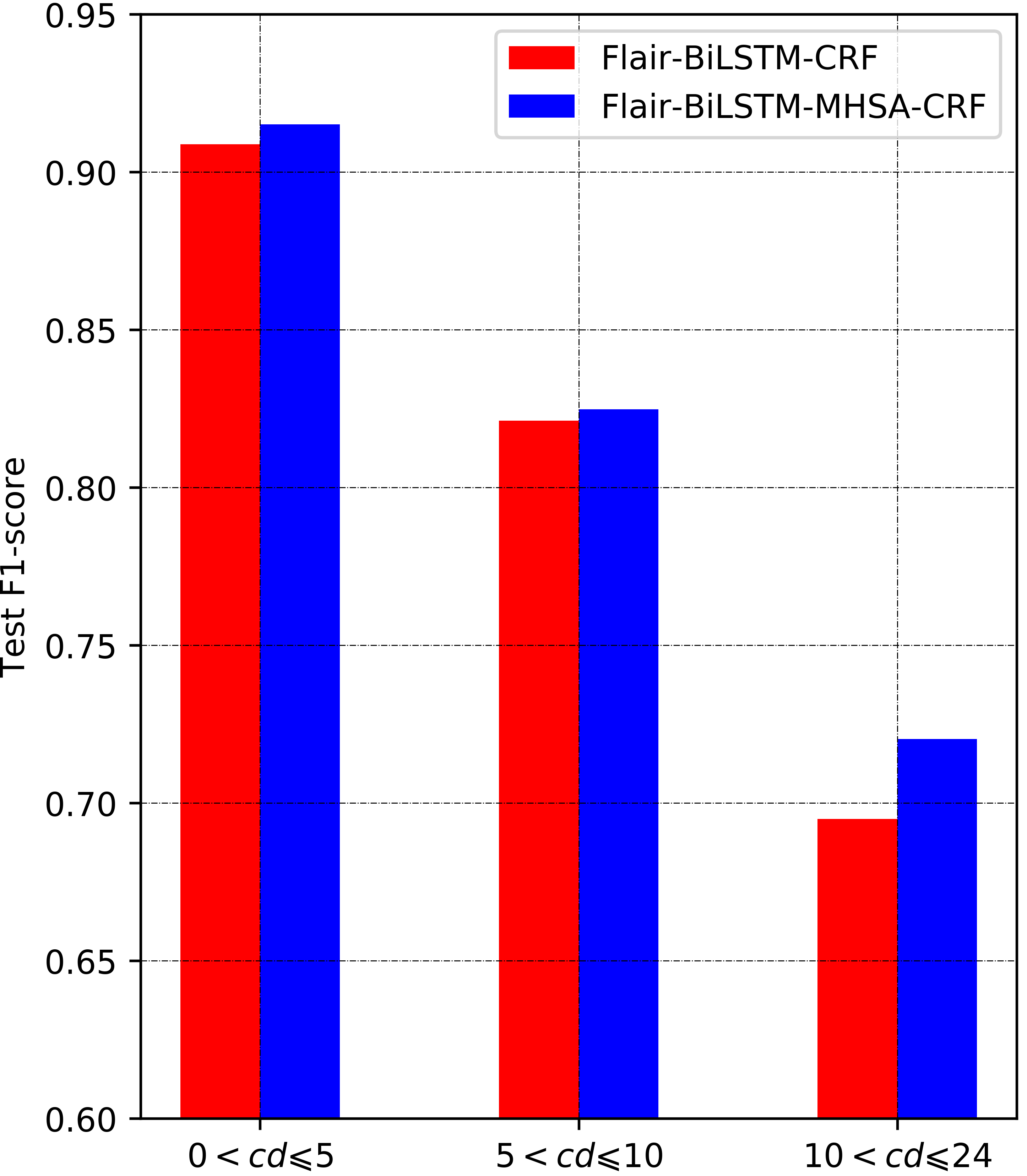}} 
    \subfloat[Group 3]{\label{fig11c}\includegraphics[scale=0.365]{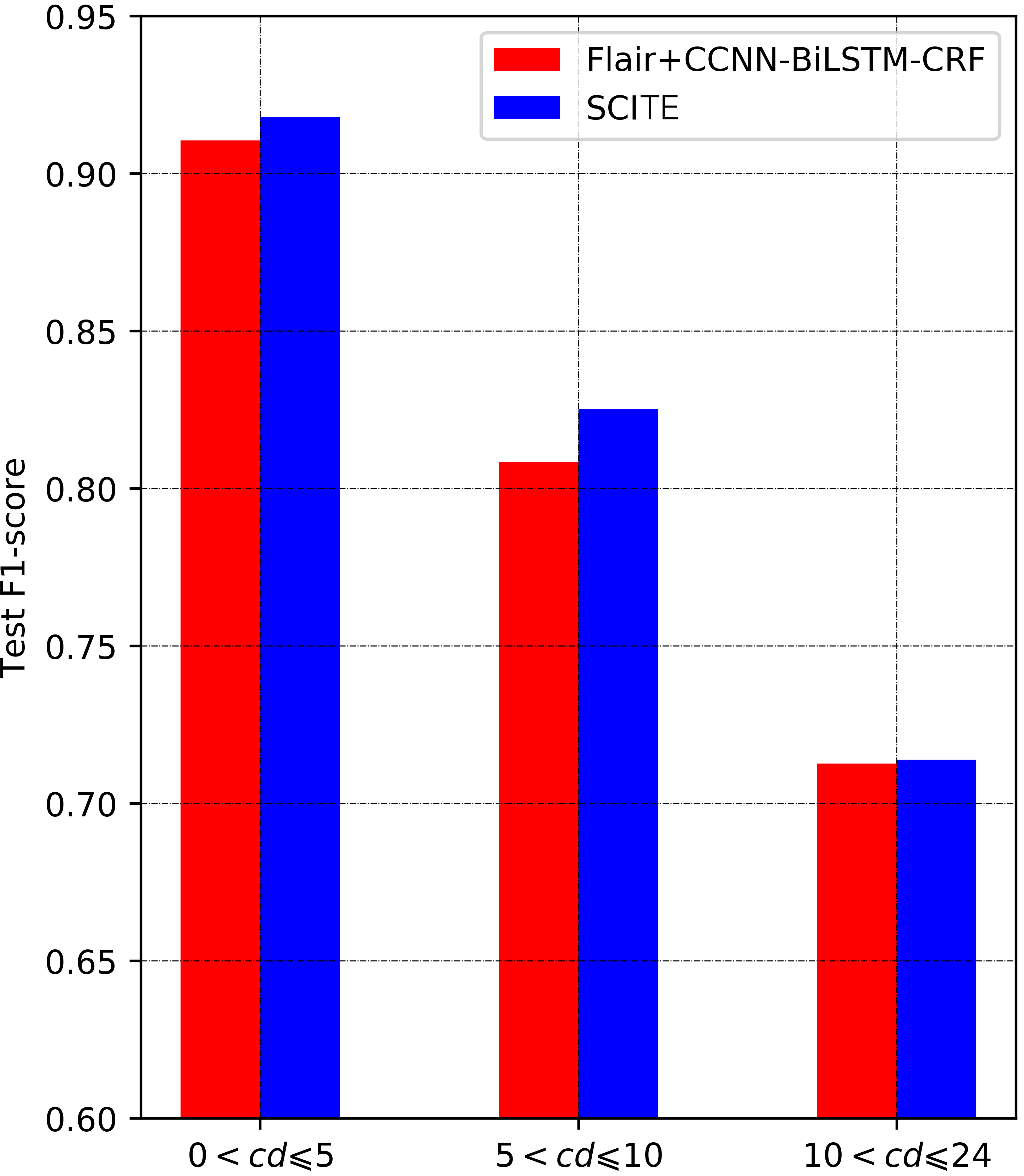}}
    \caption{Comparison of the F1-score, in terms of causality distance ($cd$). We divide the causal triplets in the test set into three parts according to their $cd$: $0<cd\leq 5$, $5<cd\leq 10$, and $10 <cd \leq 24$ (the maximum $cd$ in the test set is 24), the ratio is approximately 2:2:1.}
    \label{fig11}
\end{figure}

As shown in Fig.~\ref{fig11}, we find that the F1-scores decrease with increasing causality distance in all three groups. This validates our assumption that the long-range dependency between cause and effect creates difficulty in causality extraction. In addition, we also see that the performance of models with MHSA is better than that of models without MHSA in arbitrary causality distance, which indicates that the MHSA mechanism plays a crucial role in efficiently enhancing the association between cause and effect. In particular, MHSA significantly improves the performance in terms of the causality distance greater than 10 compared with other cases of shorter causality distance, as shown in Fig.~\ref{fig11a} and Fig.~\ref{fig11b}.

\subsection{Case Study \label{sec4.3}}

\begin{table}
\scriptsize
\setlength{\tabcolsep}{2.5pt}
\center
\caption{Results of causality extraction, where ``C'' represents ``cause'', and ``E'' represents ``effect''. We italicize \textit{\textcolor{blue}{correct}} results and highlight the \textbf{\textcolor{red}{incorrect}} results in bold. \label{tab6}}
\begin{tabular}{cc}
\toprule
Sentence 1    & \begin{tabular}[c]{@{}c@{}}\textbf{{[}The damages{]}} caused by \textbf{{[}mudslides{]}}, \textbf{{[}tremors{]}}, \textbf{{[}subsidence{]}}, \\ \textbf{{[}superficial or underground water{]}} were verified, \\ as well as \textbf{{[}swelling clay soils{]}}.\end{tabular}                                                                                                                                                                          \\ \hline
True Triplets & \begin{tabular}[c]{@{}c@{}}  \textit{\textcolor{blue}{\{{[}mudslides{]}, C-E, {[}The damages{]}\}}},\\ \textit{\textcolor{blue}{\{{[}tremors{]}, C-E, {[}The damages{]}\}}},\\ \textit{\textcolor{blue}{\{{[}subsidence{]}, C-E, {[}The damages{]}\}}},\\ \textit{\textcolor{blue}{\{{[}superficial or underground water{]}, C-E, {[}The damages{]}\}}},\\ \textit{\textcolor{blue}{\{{[}swelling clay soils{]}, C-E, {[}The damages{]}\}}}\end{tabular} \\ \hline
SCITE         & \begin{tabular}[c]{@{}c@{}} \textit{\textcolor{blue}{\{{[}mudslides{]}, C-E, {[}The damages{]}\}}},\\ \textit{\textcolor{blue}{\{{[}tremors{]}, C-E, {[}The damages{]}\}}},\\ \textit{\textcolor{blue}{\{{[}subsidence{]}, C-E, {[}The damages{]}\}}},\\ \textbf{\textcolor{red}{\{{[}superficial{]}, C-E, {[}The damages{]}\}}},\\ \textbf{\textcolor{red}{\{{[}underground water{]}, C-E, {[}The damages{]}\}}}\end{tabular} \\ \hline
Flair-BiLSTM-CRF         & \begin{tabular}[c]{@{}c@{}} \textit{\textcolor{blue}{\{{[}mudslides{]}, C-E, {[}The damages{]}\}}},\\ \textit{\textcolor{blue}{\{{[}tremors{]}, C-E, {[}The damages{]}\}}},\\ \textit{\textcolor{blue}{\{{[}subsidence{]}, C-E, {[}The damages{]}\}}}, \\ \textbf{\textcolor{red}{\{{[}underground water{]}, C-E, {[}The damages{]}\}}} \\ \textbf{\textcolor{red}{None}}\end{tabular}                                                                                                                          \\ \hline
BiLSTM-CRF         & \begin{tabular}[c]{@{}c@{}} \textit{\textcolor{blue}{\{{[}mudslides{]}, C-E, {[}The damages{]}\}}},\\ \textit{\textcolor{blue}{\{{[}tremors{]}, C-E, {[}The damages{]}\}}},\\ \textit{\textcolor{blue}{\{{[}subsidence{]}, C-E, {[}The damages{]}\}}}, \\ \textbf{\textcolor{red}{\{{[}underground water{]}, C-E, {[}The damages{]}\}}} \\ \textbf{\textcolor{red}{None}}\end{tabular}                                                                                                                          \\ \hline \hline
Sentence 2    & \begin{tabular}[c]{@{}c@{}}This year's Nobel Laureates in Physiology or Medicine \\ made the remarkable and unexpected discovery \\ that \textbf{{[}inflammation{]}} in the stomach as well as \textbf{{[}ulceration{]}} \\ of the stomach or duodenum is the result of \textbf{{[}an infection{]}} \\ of the stomach caused by \textbf{{[}the bacterium Helicobacter pylori{]}}.\end{tabular}   \\ \hline
True Triplets & \begin{tabular}[c]{@{}c@{}} \textit{\textcolor{blue}{\{{[}an infection{]}, C-E, {[}inflammation{]}\}}},\\ \textit{\textcolor{blue}{\{{[}an infection{]}, C-E, {[}ulceration{]}\}}},\\ \textit{\textcolor{blue}{\{{[}the bacterium Helicobacter pylori{]}, C-E, {[}an infection{]}\}}}\end{tabular}                                                                                                                  \\ \hline
SCITE         & \begin{tabular}[c]{@{}c@{}} \textit{\textcolor{blue}{\{{[}an infection{]}, C-E, {[}inflammation{]}\}}},\\ \textit{\textcolor{blue}{\{{[}an infection{]}, C-E, {[}ulceration{]}\}}},\\ \textit{\textcolor{blue}{\{{[}the bacterium Helicobacter pylori{]}, C-E, {[}an infection{]}\}}}\end{tabular}                                                                                                                 \\ \hline
Flair-BiLSTM-CRF         & \begin{tabular}[c]{@{}c@{}} \scriptsize{\textbf{\textcolor{red}{\{{[}the bacterium Helicobacter pylori{]}, C-E, {[}inflammation{]}\}}}},\\ \scriptsize{\textbf{\textcolor{red}{\{{[}the bacterium Helicobacter pylori{]}, C-E, {[}ulceration{]}\}}}}, \\ \textit{\textcolor{blue}{\{{[}the bacterium Helicobacter pylori{]}, C-E, {[}an infection{]}\}}}\end{tabular}                                                                                                                 
 \\ \hline
BiLSTM-CRF         & \begin{tabular}[c]{@{}c@{}} \begin{tabular}[c]{@{}l@{}}\textbf{\textcolor{red}{\{{[}the bacterium Helicobacter pylori{]}, C-E,}} \\ \textbf{\textcolor{red}{{[}the remarkable and unexpected discovery{]}\}}}, \end{tabular} \\ \textbf{\textcolor{red}{None}},\\ \textit{\textcolor{blue}{\{{[}the bacterium Helicobacter pylori{]}, C-E, {[}an infection{]}\}}}\end{tabular}                                                                                                                 
 \\ \bottomrule
\end{tabular}
\end{table}

In Table~\ref{tab6}, we list two representative examples to show the advantages and disadvantages of our proposed model. For each case, we show the input sentence and causal triplets contained in the sentence in the first and second row. The remaining rows show the extracted causal triplets of different models \footref{ft5}.

Sentence 1 is the case of simple causality (see Section~\ref{sec2.2.1}), in which five causal triplets are waiting for models to extract. We observed that neither the SCITE model nor the other two baseline models could obtain all causal triplets correctly. It seems to be difficult for the models to learn that ``superficial or underground water'' is a complete semantic unit and the phrase ``as well as'' may play a key role in 
connecting two causal components. The reason may be that the sequence tagging models based on our causality tagging scheme require slightly more training data to learn these kinds of causality expression patterns.

Sentence 2 is the case of complex causality (see Section~\ref{sec2.2.2}), in which there is an embedded causality and therefore brings difficulty and ambiguity to the causality learning of the model. In this example, only the SCITE can capture all the dependencies between cause and effect and thus precisely extract all three causal triplets when compared with other models.

\section{Related Works}

In this section, we briefly introduce causality extraction techniques proposed by other researchers, which fall into three categories: 1) approaches that employ pattern matching only, 2) techniques based on the combination of patterns and machine learning, and 3) methods based on deep learning techniques.

\subsection{Pattern-Based Methods \label{sec5.1}}

Pattern-based methods extract causality through pattern matching using semantic features, lexicon-syntactic features, and self-constructed constraints. For example, \citet{khoo1998automatic} extracted causal knowledge from the Wall Street Journal using linguistic clues and pattern matching. In the domain of the medical abstract, \citet{khoo2000extracting} used graphical patterns to extract causal knowledge from a medical database. \citet{girju2002text} extracted causal relations using the syntactic pattern ``NP1 causal-verb NP2'' with causative verbs and then employed semantic constraints to classify candidates as causal or noncausal. \citet{ittoo2011extracting} proposed a causal pair extraction method based on part-of-speech, syntactic analysis, and causality templates. In their work, causality templates were first extracted using causal sentences on Wikipedia, and then they used these templates to extract causal relations in other sentences.

These methods that rely solely on rules for pattern matching often have poor cross-domain applicability and may require extensive domain knowledge in solving problems in a particular area, as well as formulating rules that consume significant amounts of time and effort.

\subsection{Methods Based on the Combination of Patterns and Machine Learning \label{sec5.2}}

Methods based on the combination of patterns and machine learning techniques mainly treat this task in a pipeline manner. They first extract candidate phrase (or entity, event) pairs that may have causal relations according to templates or some clue words and then classify the candidate causal pairs according to some statistical features or semantic features and grammatical features to filter noncausal pairs. \citet{girju2003automatic} used constraints based on causality trigger words to extract causal relations in English texts and used the C4.5 decision tree to perform classification. \citet{sorgente2013automatic} used predefined templates to extract candidate causal pairs and then used Bayesian classifier and Laplace smoothing to filter noncausal pairs. \citet{zhao2016event} proposed a new feature called “causal connectives” by computing the similarity of the syntactic dependency structure of sentences. They run a partial parser to extract candidate noun phrases first and then classified the candidate causal pairs using the restricted hidden naive Bayes learning algorithm in combination with other features, but their method cannot discriminate the causes from the effects. \citet{luo2016commonsense} extracted cause-effect terms from large-scale web text corpora using causal cues and then used a new statistical metric-based pointwise mutual information (PMI) to measure causal strength between any two pieces of short texts.

The above methods divide causality extraction into two subtasks: candidate causal pair extraction and relation classification (filtering noncausal pairs). The results of candidate causal pair extraction may affect the performance of relation classification and generate cascading errors. These methods often require considerable human effort and time in feature engineering, relying heavily on the manual selection of textual features, and the hand-selected features are relatively too simple to capture the in-depth semantic information of the context.

\subsection{Methods Based on Deep Learning Techniques \label{sec5.3}}

Due to the powerful representation learning capabilities of deep neural networks that can effectively capture implicit and ambiguous causal relations, the adoption of deep learning techniques for causality extraction has become a popular choice for researchers in recent years. \citet{de2017causal} used CNN to classify causal relations in the text. \citet{kruengkrai2017improving} used multicolumn CNN with the background knowledge extracted from noisy texts to classify such commonsense causalities as ``smoke cigarettes" $\rightarrow$ ``die or lung cancer''. Similarly, \citet{Li_2019} proposed a knowledge-oriented CNN that incorporates prior knowledge from lexical knowledge bases for causal relation classification. \citet{martinez2017neural} proposed an LSTM-based model only fed with word embeddings for the task of causality classification. In addition to classifying causality from a common sense reasoning standpoint, \citet{dasgupta2018automatic} and \citet{dunietz2018deepcx} also identified the linguistic expressions of causality in the text from a linguistic point of view through deep LSTM-based models.

The main differences between our proposed method and the above methods based on deep learning techniques can be summarized as follows:

\begin{itemize}
\item Our method aims to automatically extract such common sense causal triplets as c in the text (Fig.~\ref{fig1}), not only to classify causal relations or to identify the linguistic expressions of causality.
\item Our method can easily handle multiple causal triplets and embedded causality in the same sentence (Section \ref{sec2.1} and Section \ref{sec2.2}) without having to divide the sentences into subsentences that contain only one instance of causality and thus generate cascading errors as in \citet{dasgupta2018automatic}.
\end{itemize}

\section{Conclusion}

In this paper, we formulate causality extraction as a sequence tagging problem and deliver a self-attentive BiLSTM-CRF-based solution for
the causality extraction. In particular, we propose SCITE to extract causality in natural language text based on our causality tagging scheme. To alleviate the problem of data insufficiency, we transfer the Flair embeddings trained from a large corpus into our task. In addition, we introduce the multihead self-attention mechanism to learn the dependencies between cause and effect. Experimental results demonstrate the effectiveness of our proposed method. However, the performance of SCITE is still limited to some extent by the insufficiency of high-quality annotated data (Section~\ref{sec4.3}).

In future work, we will attempt to solve this problem as follows:

\begin{enumerate}
\item Develop annotated datasets from multiple sources based on existing datasets and our causality tagging scheme.
\item Combine our method with distant supervision \cite{mintz2009distant} and reinforcement learning \cite{sutton2018reinforcement} to achieve better performance without having to build a high-quality annotated corpus for causality extraction.
\end{enumerate}

\section{Acknowledgments}

This research is partially supported by the National Natural Science Foundation of China (No. U1711263).

\section*{References}

\bibliography{reference}

\end{document}